\documentclass[conference]{IEEEtran}
\IEEEoverridecommandlockouts

\usepackage{cite}
\usepackage[T1]{fontenc}
\usepackage{type1cm}
\usepackage[pdfversion=1.7]{hyperref} 
\usepackage{amsmath,amssymb,amsfonts}
\usepackage{newtxtext,newtxmath}
\usepackage{algorithmic}
\usepackage{graphicx}
\usepackage{textcomp}
\usepackage{xcolor}
\usepackage{algorithm}
\usepackage{algorithmic}
\usepackage{multirow}
\usepackage{booktabs}
\usepackage{subcaption}
\usepackage{array}
\def\BibTeX{{\rm B\kern-.05em{\sc i\kern-.025em b}\kern-.08em
    T\kern-.1667em\lower.7ex\hbox{E}\kern-.125emX}}
\begin{document}

\title{Multi-Scale Global-Instance Prompt Tuning for Continual Test-time Adaptation in \\ Medical Image Segmentation\\
}

\author{
\IEEEauthorblockN{
Lingrui Li$^{1}$, 
Yanfeng Zhou$^{2}$, 
Nan Pu$^{3*}$,
Xin Chen$^{1*}$, 
Zhun Zhong$^{1,4}$
}
\IEEEauthorblockA{
$^{1}$School of Computer Science, University of Nottingham, UK, $^{2}$School of Artificial Intelligence, Shenzhen University, China, \\
$^{3}$Department of Information Engineering and Computer Science, University of Trento, Italy,  \\ $^{4}$School of Computer Science and Information Engineering, Hefei University of Technology, China
}
\thanks{*Corresponding author.}
\thanks{Author’s Accepted Manuscript. Released under the Creative Commons license: Attribution 4.0 International (CC BY 4.0) https://creativecommons.org/licenses/by/4.0/}
}

\maketitle

\begin{abstract}
Distribution shift is a common challenge in medical images obtained from different clinical centers, significantly hindering the deployment of pre-trained semantic segmentation models in real-world applications across multiple domains. Continual Test-Time Adaptation (CTTA) has emerged as a promising approach to address cross-domain distribution shifts during continually evolving target domains. 
{\color{black}{Most existing CTTA methods rely on incrementally updating model parameters, which inevitably suffer from error accumulation and catastrophic forgetting, especially in long-term adaptation. Recent prompt-tuning-based works have shown potential to mitigate the two issues above by updating only visual prompts. While these approaches have demonstrated promising performance, several limitations remain: 1) lacking multi-scale prompt diversity, 2) inadequate incorporation of instance-specific knowledge, and 3) risk of privacy leakage.}}
To overcome these limitations, we propose \textbf{M}ulti-scale \textbf{G}lobal-\textbf{I}nstance \textbf{P}rompt \textbf{T}uning (MGIPT), to enhance scale diversity of prompts as well as capture both global- and instance-level knowledge for robust CTTA. Specifically, MGIPT consists of an Adaptive-scale Instance Prompt (AIP) and a Multi-scale Global-level Prompt (MGP). AIP dynamically learns lightweight and instance-specific prompts to mitigate error accumulation with adaptive optimal-scale selection mechanism. MGP captures domain-level knowledge across different scales to ensure robust adaptation with anti-forgetting capabilities. These complementary components are combined through a weighted ensemble approach, enabling effective dual-level adaptation that integrates both global and local information.
Extensive experiments on medical image segmentation benchmarks (five optic disc/cup datasets and four polyp datasets) demonstrate that our MGIPT outperforms state-of-the-art methods, achieving robust adaptation across continually changing target domains. Notably, our MGIPT exhibits particularly strong performance in long-term CTTA scenarios, showing great anti-forgetting ability. 
\end{abstract}


\begin{figure}[htbp]
    \centering
     \includegraphics[width=\linewidth]{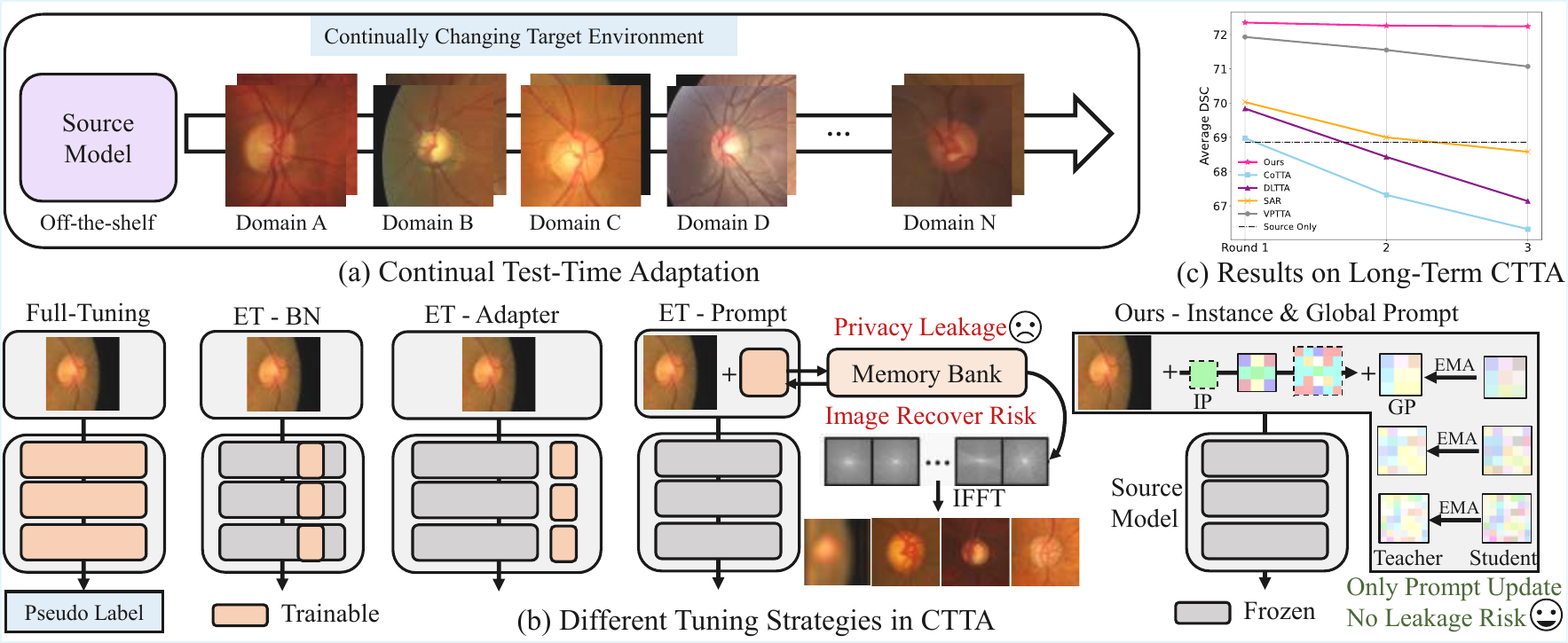}
\caption{(a) The goal of continual test-time adaptation (CTTA) is to adapt a pre-trained source model to handle data in dynamically changing target domains. (b) Overview of tuning strategies in CTTA. Fully-Tuning often causes catastrophic forgetting due to frequent parameter updates and error accumulation due to noisy pseudo labels; Efficient-Tuning (ET) on Batch-Normalization (BN) layers and ET with adapters struggle with balancing trade-offs of inadequate adaptation or error accumulation; ET with single-scale global prompts leads to suboptimal individual instance adaptation and privacy leakage. Our method addresses error accumulation and forgetting by leveraging dual-level prompts with multiple scales and robust parameter updating strategies while reducing the risk of privacy leakage. (c) Performance comparison on OD/OC datasets over 3 continual rounds. Our approach demonstrates superior performance over previous methods.}
\label{fig1-motivation}
\end{figure}

\section{Introduction}
\label{sec:intro}
Deep image segmentation models trained on data from a single clinical setting often fail when applied to images from other hospitals or clinical centers, even when dealing with the same organ or cell types~\cite{survey,udamed}. These failures arise from data distribution shifts caused by differences in imaging protocols, devices, and patient demographics, posing a significant barrier to the deployment of robust models in real-world healthcare environments.

Test-Time Adaptation (TTA) methods~\cite{TENT} have emerged as practical solutions, enabling models to adapt to new data distributions without requiring access to source data. Unlike traditional Unsupervised Domain Adaptation (UDA), which demands complete access to source dataset and entire target datasets~\cite{lmisa}, TTA adapts using only sequentially presented test samples without retaining previously processed data, making it more suitable for real-world scenarios. Mainstream TTA methods often leverage self-supervised tasks - such as entropy minimization and self-training to define auxiliary losses and update model parameters~\cite{TENT,robust}. While these TTA methods effectively reduce domain gaps, they often assume static target distributions, which does not align with changing scenarios when data evolves over time. 

To meet the practical requirements, Continual Test-Time Adaptation (CTTA)~\cite{cotta} has been proposed to handle dynamic target domains, as illustrated in Fig.~\ref{fig1-motivation} (a).
CTTA faces two critical challenges: error accumulation and catastrophic forgetting. The former occurs when noisy or unreliable self-supervised tasks amplify inaccuracies over time~\cite{vida}. The latter arises when updates to the model overwrite prior knowledge, degrading performance on previous domains~\cite{sar, cotta, vptta}. Existing methods attempt to address these issues through loss optimization~\cite{domainadaptor} for improving supervision during adaptation, regularization and model-resetting strategies for anti-forgetting~\cite{sar, cotta}. However, these approaches struggle to balance adaptation and stability~\cite{vptta}: overly conservative parameter updates hinder adaptation, while excessive updates exacerbate forgetting and error accumulation. This dilemma further limits the comprehensive capability of CTTA.

To achieve a better trade-off between adaptation and stability, recent CTTA methods~\cite{vptta, vpt3} utilize prompt tuning technique~\cite{vpt} to update only visual prompts, rather than modifying model parameters. While these prompt-based methods~\cite{vptta, vpt3} show effectiveness in reducing error accumulation and forgetting, they still suffer from two main limitations. \textbf{1)} As for the type of prompts, they neglect instance-level knowledge yet learn only global prompts, which fails to capture unique characteristics of individual samples~\cite{vpt3,vptta}. This constraint leads to suboptimal optimization for individual samples and introduces errors during adaptation. 
\textbf{2)} As for the design of global prompts, they employ single-scale prompts to capture domain knowledge. This approach lacks the flexibility to represent diverse visual characteristics of multiple scales across domains, resulting in incomplete global knowledge acquisition. In addition, since each image has its distinct optimal prompt scale due to individual differences, applying the same scale for all images can cause suboptimal optimization. 
\textbf{3)} The memory bank in VPTTA\cite{vptta} introduces additional privacy concerns by continuously storing information from previously seen images. Although only the low-frequency components are preserved - rather than the raw images themselves - these components still contain substantial visual information. As illustrated in Fig.~1(b), a simple inverse Fourier transform is sufficient to reconstruct most of the original image content. This vulnerability significantly undermines the trustworthiness of VPTTA, particularly in biomedical applications where protecting sensitive patient data is critical.

To overcome these limitations, we propose Multi-scale Global-Instance Prompt Tuning (MGIPT), a novel framework that integrates multi-scale global prompts with progressively adaptive instance-level prompts for robust CTTA. 
Our approach consists of two complementary components. \textbf{1)} Adaptive-scale Instance Prompts (AIP), which mitigates error accumulation by adapting at the instance level with progressive optimization, ensuring high specificity and less error propagation. \textbf{2)} Multi-scale Global Prompts (MGP), which captures various domain-level knowledge across different scales, ensuring robust adaptation with anti-forgetting. 
By combining these two mechanisms, MGIPT achieves a balanced adaptation that maintains global knowledge while accommodating instance-specific variations, enabling reliable performance in continually evolving environments. Our contributions are summarized as follows:

\begin{itemize}
\item We propose MGIPT, a novel CTTA framework that captures diverse global and instance information through complementary prompt tuning mechanisms, effectively solving error accumulation and catastrophic forgetting.

\item We design AIP, a progressive instance-level prompt tuning mechanism that addresses instance-specific shifts while minimizing error accumulation through adaptive optimal scale selection. 

\item We introduce MGP, a multi-scale global prompt module that captures domain and continual knowledge at different granularities, enabling adaptation to diverse and evolving distributions, while reducing potential privacy concerns. 

\item Extensive experiments on medical image segmentation datasets demonstrate that MGIPT outperforms state-of-the-art methods across continually changing target domains, especially in long-term CTTA settings (Fig. 1 (c))
\end{itemize}

\section{Related Work}
\label{sec:relatedwork}

\paragraph{Continual Test-Time Adaptation}
TTA~\cite{survey,TENT} aims to bridge domain gaps when deploying pre-trained source domain models on unseen target domains by adapting to unlabeled test data during inference. Common TTA methods leverage unsupervised objectives, such as entropy minimization and self-supervised tasks, to align distributions~\cite{TENT, sar}, or pseudo-labels to drive model adaptation~\cite{robust}. Some methods also modify Batch Normalization (BN) statistics to reduce domain shifts~\cite{sita, dua}. While effective for static target domains, these approaches struggle in dynamic scenarios where distributions evolve over time. In addition, some of the methods require customized pretraining stages or additional pretrained models, diverging from the goal of performing flexible adaptation solely at test time~\cite{robust}.

CTTA extends TTA to dynamic environments, addressing the challenge of adapting models to continually changing target domains~\cite{cotta}. Techniques such as self-training, augmentation-averaged pseudo-labels, and stochastic weight resetting have been proposed to mitigate error accumulation and catastrophic forgetting~\cite{sar, vida, cotta}. 
Several parameter-efficient CTTA methods~\cite{vptta, vida} aim to overcome catastrophic forgetting and reduce computational overhead. However, these methods often require resetting model parameters and struggle with insufficient or excessive updating of the model parameters which are still impacted by forgetting and error accumulation, or maintaining memory banks that raise concerns about privacy~\cite{vida}. \textit{Unlike these approaches, our method achieves robust long-term adaptation by freezing the pretrained model and introducing dual-level multi-scale prompt tuning.} 

\paragraph{Prompt Learning}
Prompt learning originated in Natural Language Processing (NLP) as a method to enhance downstream task performance by introducing task-specific text instructions for pre-trained language models~\cite{promptnlp}. Its success in NLP has inspired its adaptation to computer vision, where visual prompts have been shown to improve generalization~\cite{vpt}. For example, visual prompts~\cite{vpt3} at the pixel level have been shown to explicitly reduce domain discrepancies, while other approaches~\cite{pass, prosfda} utilizes low-frequency prompts for domain adaptation. 
Despite these advancements, many of these approaches~\cite{vpt, pass, prosfda} often assume static target domains, which is not effective for continual adaptation across evolving data. To address this drawback, the most related work~\cite{vptta} proposes a VPTTA method, which enables prompt-based adaptation methods to handle continually evolving data distributions.

\textit{Compared with VPTTA~\cite{vptta}, our MGIPT has the following three advantages:} 1) Unlike VPTTA~\cite{vptta} that maintains only single-scale frequency prompts, our MGIPT proposes both adaptive and multiple frequency-scale prompt tuning, which improves generalization ability and flexible and comprehensive adaptation. 2) VPTTA relies on only domain-level global prompts, which ignores instance-level knowledge~\cite{vptta} and thus leads to suboptimal optimization for individual samples. In contrast, MGIPT jointly learns global-instance prompts to capture shareable domain-level knowledge while complimenting personalized instance knowledge. 3) VPTTA requires an additional memory bank~\cite{vptta} to store historical prompts at fixed scales, which accumulates noise over long term, leading to residual error propagation. Furthermore, the memory bank retains previous image data, posing a privacy risk as original images can be mostly recovered. MGIPT employs a memory-free Teacher-student update that eliminates the residual error accumulation issues while enhancing privacy preservation.

\begin{figure*}[!t]
    \centering
    \begin{subfigure}[b]{0.68\textwidth}  
        \centering
        \includegraphics[width=\textwidth]{fig2-1.png}
        \label{fig:subfig2-1}
    \end{subfigure}
    \hfill  
    \begin{subfigure}[b]{0.31\textwidth}  
        \centering
        \includegraphics[width=\textwidth]{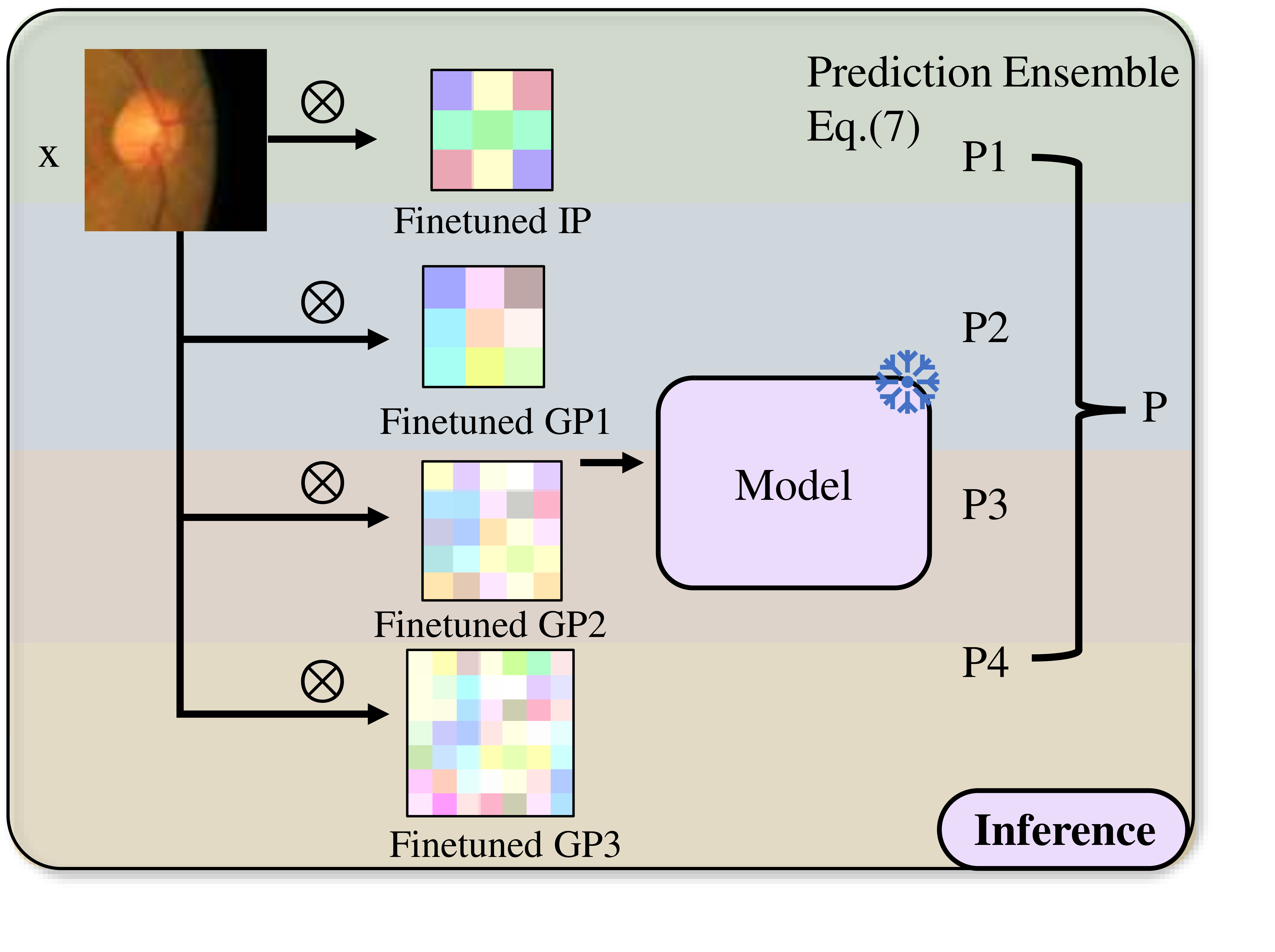}  
        \label{fig:subfig2-3}
    \end{subfigure}
    \vspace{-25pt}  
    \caption{The framework of MGIPT. This framework leverages instance-specific and global knowledge to tackle the error accumulation and forgetting challenges during CTTA. MGIPT comprises: (1) Low-Frequency Prompts: trainable pixels initialized as ones, optimized using Batch Normalization (BN) alignment loss. 
   (2) Adaptive-scale Instance Prompt: input images are transformed to the frequency domain via Fast Fourier Transform (FFT), combined with instance-level prompts (followed by inverse FFT), and optimized using progressive prompt tuning (PPT) with an early stopping (ES) strategy for optimal scale selection.
   (3) Multi-scale Global Prompt: captures domain-level knowledge across varying receptive fields using multi-scale prompts, updated with a teacher-student mechanism and Exponential Moving Average (EMA).
   (4) Inference: fine-tuned instance and global prompts are combined to adapt the input image, producing the final output.}
\label{fig2}
\end{figure*}

\vspace{-1pt}
\section{Methodology}
\par\noindent\textbf{Task Definition.} We tackle the challenge of adapting a pre-trained source model $f_s: X \rightarrow Y$ to target data from continually evolving, non-stationary environments at test time - without accessing to labeled source or target data. Let the source dataset be $ \mathcal{D}_s = \{({x}_i^s, y_i^s)\} $ where ${x}_i^s$ is the $i$-th source image and $y_i^s$ is its label, and the target data stream at time step $ t $ be $ \mathcal{D}_t = \{{x}_i^t\}$, where ${x}_i^t$ is the $i$-th target image of size $H \times W$, with $\mathcal{D}_t$ changing over time. The goal of CTTA is to adapt the source model effectively across these evolving target domains over $ T $ sequential rounds ($ t = 1, \ldots, T $).

\par\noindent\textbf{Framework Overview.}
To achieve the above goal in CTTA, our proposed MGIPT framework utilizes low-frequency prompt tuning with two complementary components.
\textit{Adaptive-scale Instance Prompts (AIP)}: designed to address instance-specific distribution shifts via progressive adaptations and optimal scale selection.
\textit{Multi-scale Global Prompts (MGP)}: designed to capture domain-level knowledge across multiple frequency scales, updated through a teacher-student strategy. 
The two components are dynamically combined using confidence-based weights to balance global and local adaptation. An overview of MGIPT is shown in Fig.~\ref{fig2}, and each component is detailed below.

\subsection{Low-Frequency Prompt Design}
\par\noindent\textbf{Prompt Mechanism and Implementation.} Previous works~\cite{fda, prosfda} have shown that low-frequency components are strongly associated with style textures, which are the primary factor in distribution shifts. We therefore employ low-frequency visual prompts to adapt the amplitude spectrum of different input images at inference time.
Let \(\mathcal{F}(\cdot)\) and \(\mathcal{F}^{-1}(\cdot)\) denote the Fast Fourier Transform (FFT) and Inverse FFT (IFFT), respectively. The magnitude and phase components are represented as \(\mathcal{F}^M(\cdot)\) and \(\mathcal{F}^P(\cdot)\). All prompts are initialized to ones. For a prompt \( P_i \in \mathbb{R}^{(H_P) \times (W_P) \times C} \), the adapted image \(\hat{{X}}_i^t\) is given by:
\begin{equation}
\hat{X}_{i}^{t} = \mathcal
{F}^{-1}([\text{OnePad}(P_{i}) \odot \mathcal{F}^{M}(X_{i}^{t}), \mathcal{F}^{P}(X_{i}^{t})]),
\label{eqn:fftinput}
\end{equation}
where \(\odot\) denotes element-wise multiplication. \(\text{OnePad}(\cdot)\) means padding \( P_i \) with surrounding ones to match the spatial dimensions of \(\mathcal{F}^M({X}_i^t)\), whose size is $ H\times W$. This design ensures that the prompt $P_i$ remains lightweight and focuses on the low-frequency component.

\par\noindent\textbf{Prompt Training.}
In this paper, we train the Low-frequency prompt $P^{ip}_i$ with Batch normalization (BN) alignment loss of the BN statistics (mean \(\mu_s\) and standard deviation \(\Sigma_s\)) stored in the source model and BN statistics calculated on the features of the test image (\(\mu_t, \Sigma_t\))~\cite{adabn}:
\begin{equation}
\mathcal{L}_{align} = \frac{1}{j} \sum |\mu_s^j - \mu_t^j| + |\Sigma_s^j - \Sigma_t^j|,
\label{eqn:loss_s}
\end{equation}
where  \( j \) indexes the BN layers in \( f_s \).
This BN loss is used for both instance-level and global-level prompt learning. 

\subsection{Adaptive-scale Instance-level Prompt} 
Distribution shifts affect each image differently based on its unique characteristics, suggesting that optimal prompt sizes should vary between instances. Applying a single scale across all images can lead to suboptimal adaptation and introduce errors during continual adaptation. Therefore, determining the optimal scale for each instance becomes crucial for effective adaptation. 
\par\noindent\textbf{Progressive Prompt Tuning.}
To capture instance-specific information, MGIPT iteratively enlarges the prompt region across $k$ steps. At iteration $k+1$, the instance prompt $P_{k+1}^{ip}$ is updated as:
\begin{equation}
P_{k+1}^{ip} = \text{OnePad}(P_k^{ip}),
\label{eqn:pk+1}
\end{equation}
where $P_k^{ip}$ is the fine-tuned instance prompt from the previous iteration, and size of $P_{k+1}^{ip}$ is ${(H_P+2) \times (W_P+2) \times C}$, starting with initial $H_P$ and $W_P$ size of 1.  By fine-tuning the region of $P_{k+1}^{ip}$ that excludes the previous $P_{k}^{ip}$ region (frozen), MGIPT prevents overfitting while capturing fine-grained instance variations.

\textit{Instance prompts are designed to dynamically adapt to sample-specific variations. This enhances the model's ability to handle intra-domain shifts and reduces catastrophic forgetting of global knowledge during continual adaptation. This mechanism also minimizes error accumulation caused by the misalignment of sample-specific distributions.}

\par\noindent\textbf{Early Stopping for Adaptive Prompt Scale.}
While the progressive approach allows for continuous expansion, each instance has a unique optimal scale, necessitating an automatic termination criterion. We propose a consistency-based measure to determine when to stop the iterative process.
To identify the optimal scale for each instance prompt (IP) and prevent overfitting, we monitor the consistency between predictions on the original and augmented inputs. We employ DiceLoss~\cite{diceloss} as a quantitative measure to determine when to terminate the progressive prompt iteration:
\begin{equation}
\mathcal{M}_{\text{cons}} = \text{DiceLoss}(PL, PL_{\text{aug}}),
\label{eqn:l_cons}
\end{equation}
where pseudo labels $ {PL} = \text{Sigmoid}(p) > 0.5$, and ${PL}_{aug} = \text{Sigmoid}(p_{aug}) > 0.5$ derived from the model predictions $p$ and ${p}_{aug}$, respectively. For each input, we generate two augmented variants by applying color jitter transformations~\cite{aug} with modified brightness and contrast parameters. The final augmented prediction $p_{aug}$ is obtained by averaging predictions across augmented images.

\textit{The instance prompt iteration process automatically terminates when $M_{cons}$ no longer improves, indicating that the instance prompt has reached its optimal configuration. This stopping criterion ensures robust prompts without introducing noisy updates during the adaptation process.}

\subsection{Multi-scale Global-level Prompt Tuning}
\par\noindent\textbf{Multi-scale Prompt Ensembles.}
Earlier prompt-based approaches~\cite{vptta,vpt3} rely on single-scale prompts for adaptation. However, single-scale prompts are limited in their ability to capture global knowledge, as they focus on a fixed frequency band of domain characteristics. This constraint prevents them from encompassing the diverse visual features present across multiple scales, leading to incomplete acquisition of global knowledge.
To address this limitation, we propose a multi-scale global prompt ensemble strategy to capture a more diverse spectrum of domain knowledge across varying scales. 

Global prompts (GP) are designed to capture continual knowledge and long-term domain changes. These prompts are initialized as low-frequency ones, similar to IP. This strategy utilizes global prompts of three distinct sizes (\( bs-2, bs, bs+2 \)) to capture different amplitude aspects of the original images.
While we adopt VPTTA~\cite{vptta} empirically determined optimal base size of global prompt $bs$ as our central scale, we incorporate smaller $bs-2$ and larger $bs+2$ scales to promote diverse style adaptation across frequency bands.   
This multi-scale design yields three global prompts, each contributing unique representational characteristics to the ensemble. By leveraging these varied perspectives, our approach can better capture broader domain-level patterns as shown in Table~\ref{tab:ablation-study} comparison between Row \#4 \& Row \#5, improving robustness across diverse target domains.

\par\noindent\textbf{Prompt Design with Teacher-Student Strategy.}
Unlike prior works~\cite{vptta} that use memory banks to store historical prompts for obtaining global knowledge, we adopt a teacher-student prompt updating strategy. Specifically, the student prompt $P^{sg}_i$ is optimized with the BN alignment loss of the source (\(\mu_s, \Sigma_s\)) and target (\(\mu_t, \Sigma_t\)) statistics as in Eq. (\ref{eqn:loss_s}).
The teacher prompt $P^{tg}_i$ is updated via Exponential Moving Average (EMA) mechanism:
\begin{equation}
P_i^{tg} = e P_{i-1}^{tg} + (1 - e) P_i^{sg},
\label{eqn:ema}
\end{equation}
where \( e \) is the EMA decay factor.
In conjunction with the multi-scale strategy, we have three teacher global prompts $P^{tg1}_i$, $P^{tg2}_i$, $P^{tg3}_i$ that are initialized to ones and updated independently by Eq. (\ref{eqn:ema}). This teacher-student strategy eliminates the need for storing historical prompts, thereby improving privacy preservation. Moreover, it provides a more consistent knowledge transfer mechanism that captures domain knowledge and mitigates catastrophic forgetting.
\subsection{Inference}
Batch normalization (BN) calibration is widely used in domain adaptation~\cite{dua} to adjust BN statistics for the target domain. In cases with limited target samples, such as in CTTA, directly using target statistics ($\mu_t$, $\Sigma_t^2$) can yield unreliable estimates, while relying solely on source statistics ($\mu_s$, $\Sigma_s^2$) may fail due to internal covariate shifts. 
To address this, we use a common weighted combination~\cite{sita} to update $\mu_s$ and $\Sigma_s^2$ for calibration:
\begin{equation}
\mu = \lambda \mu_s + (1 - \lambda) \mu_t, \quad \Sigma^2 = \lambda \Sigma_s^2 + (1 - \lambda) \Sigma_t^2,
\label{eqn:bnCalibra}
\end{equation}
where $\lambda \in [0,1]$ balances source and target statistics.

The proposed MGIPT framework integrates AIP and MGP as complementary modules to address fine-grained local variations and domain shifts. These two components achieve synergy during inference by dynamically combining predictions based on confidence-weighted ensembles.
Predictions from the three global prompts $p_{gp}^n$ (n=3) and from the instance prompt $p_{ip}$ are combined to produce the final output $O$ using confidence-based weights $w_{gp}$ and $w_{ip}$:
\begin{equation}
O = w_{ip} p_{ip} + \sum_n w_{gp}^n p_{gp}^n, \quad w_t = \frac{\mathcal{C}_t}{\sum \mathcal{C}_t},
\label{eqn:mspromptEnsemble}
\end{equation}
where confidence $\mathcal{C}_t = \max(\text{softmax}(p_t))$ ensures predictions are weighted by confidence.

\subsection{Discussion on Distinct Scale Optimization Strategy for AIP and MGP}
The optimization strategies for AIP and MGP are distinct due to their different objectives.
AIP: focuses on instance-specific adaptation, ensuring that the prompts can precisely capture the unique characteristics of each sample at an optimal scale. 
To identify this scale, we adopt a progressive approach: starting from a minimal scale, it incrementally refines the adaptation by iteratively expanding the receptive field, with an early stopping mechanism to prevent overfitting. 
MGP: unlike AIP, MGP addresses domain-level adaptation. Since samples within a domain can vary in their optimal scales, relying on a single scale would risk overlooking useful knowledge from other scales. To mitigate this, MGP is designed to extract information from multiple scales simultaneously. By incorporating prompts across diverse frequency range, MGP ensures a more comprehensive understanding of the global knowledge, solving the limitations of single-scale optimization.

\section{Experiments}

\subsection{Datasets and Evaluation Metrics}
We evaluate our method on two segmentation tasks: Optic Disc/Optic Cup (OD/OC) segmentation and polyp segmentation.
The OD/OC segmentation dataset~\cite{vptta} comprises five public datasets from different medical centers: Domain A (RIM-ONE-r3~\cite{RIM}), B (REFUGE~\cite{Refuge}), C (ORIGA~\cite{origa}), D (REFUGE-Validation/Test~\cite{Refuge}), and E (Drishti-GS ~\cite{Drishti}), with 159, 400, 650, 800, and 101 images, respectively. Regions of interest (ROIs) centered on the OD were cropped to $800\times800$, resized to $512\times 512$, and normalized using min-max normalization, following~\cite{vptta, prosfda}. The Dice Similarity Coefficient (DSC) was used for evaluation. The polyp segmentation dataset consists of four public datasets from different medical centers: Domain A (BKAI-IGH-NeoPolyp~\cite{polypdataset1}), B (CVC-ClinicDB~\cite{polypdataset2}), C (ETIS-LaribPolypDB~\cite{polypdataset3}), and D (Kvasir-Seg~\cite{polypdataset4}), with 1000, 612, 196, and 1000 images, respectively. Following~\cite{pranet,vptta}, images were resized to $352 \times 352$ and then normalized using ImageNet statistics. Evaluation metrics included DSC, enhanced-alignment metric ($E^{max}_{\phi}$)~\cite{enhancedmetric}, and structural similarity metric ($S_{\alpha}$)~\cite{SMmetric}.
 
\begin{table}[!t]
    \centering
    \scriptsize
    \setlength{\tabcolsep}{3pt}
    \caption{Results of ablation study on the OD/OC segmentation task. The best and second-best results in each column are highlighted in \textbf{bold} and \underline{underline}, respectively.}
    \scalebox{0.91}{
    \begin{tabular}{cc|ccc|ccc|ccccc|c}
        \toprule
        \multirow{
2}{*}{\#}&
         \multirow{
2}{*}{BN } &\multicolumn{3}{c|}{\textcolor{purple}{GP}} 
          & \multicolumn{3}{c|}{\textcolor{orange}{IP}} & \multicolumn{5}{c|}{Domain} & \multirow{2}{*}{Avg} \\
        \cmidrule(lr){3-13}
         &Calib. & Fix &EMA &MS & Fix & PPT &ES & A & B & C & D & E & DSC $\uparrow$ \\

        \midrule 
        1& &  & & &  &  &  &  64.53 & 76.06 & 71.18 & 52.67 & 64.87 & 65.86   \\
        2&\checkmark &  & & &  &  &  &  74.51 &77.12 &\underline{73.77}& 53.6& 66.28& 69.06   \\
        3&\checkmark & \checkmark & &  & &  &  & 74.59 & 77.55 & 71.96 & 57.11 & 74.61 & 71.16 \\
        4&\checkmark & \checkmark & \checkmark & &  & &  & 74.56 & 77.72 & 72.48 & 57.17 & 74.14 & 71.21 \\
        5&\checkmark & \checkmark & \checkmark & \checkmark & & &   & 74.59 & 77.94 & 72.42 & \underline{57.17} & 75.29 & 71.48 \\
        6&\checkmark & \checkmark & \checkmark & \checkmark & \checkmark & &   &  74.02&  78.51 & 71.67 & \textbf{57.46} & 75.94 & 71.52  \\
        7&\checkmark & \checkmark & \checkmark & \checkmark & & \checkmark & & \underline{74.66} & \underline{78.83} & 73.34 & 56.19 & \underline{77.42} & \underline{72.09} \\
        8&\checkmark & \checkmark & \checkmark & \checkmark & & \checkmark &\checkmark & \textbf{75.14} & \textbf{78.92} & \textbf{73.89} & 55.89 & \textbf{77.91} & \textbf{72.35} \\ 
        \bottomrule
    \end{tabular}}
    \label{tab:ablation-study}
\end{table}

\subsection{Implementation Details}
For each task, we trained the source model on each single domain (source domain) and tested it on the remaining domains (target domains) to calculate the mean values of metrics for evaluation. 
In the source-training phase, we trained a source model with ResUNet-34~\cite{resUnet} backbone, following~\cite{vptta,prosfda}, as the baseline for OD /OC segmentation task, and trained the PraNet~\cite{pranet} with a Res2Net-based~\cite{res2net} backbone for the polyp segmentation task.  
In the test-adaptation phase, instance prompts and global prompts were updated for 7 and 1 epochs, respectively. Due to the complexity of PraNet’s decoder, the prompt training loss \( L_{align} \) was calculated only for the BN layers in the encoder instead of the whole network~\cite{vptta}. Batch size of 1 was used on all experiments of our method and other competing methods. The competing methods were evaluated with identical configurations to ensure fairness. We employed the Adam optimizer with a learning rate of 0.05 and 0.01 for the OD/OC segmentation task and polyp segmentation task, respectively. 
The hyperparameters \( e \) (EMA decay weight), $\lambda$ (weight of BN calibration) and bs (base size of global prompt) were set to 0.1, 0.8, and 5 (following~\cite{vptta}) for both tasks.

\begin{table}[!t]
\centering
\small
\caption{Comparison of Dice Similarity Coefficient (DSC) on the OD/OC segmentation task across different methods and domains. The best performance is highlighted in \textbf{bold}. * denotes statistical significance (p \textless 0.05) using Wilcoxon signed-rank test~\cite{Wilcoxon} by comparing other methods to ours. ``Avg'' means ``Average''. $^{\dagger}$ denotes removing memory bank.}
\scalebox{0.78}{
\begin{tabular}{l|c|c|c|c|c|c}
\hline
Methods/ DSC $\uparrow$ & A &  B  &  C &  D  &  E  & Avg  $\uparrow$ \\ \hline
Source Only & 64.53 & 76.06 & 71.18 & 52.67 & 64.87 & 65.86 \\ \hline
TENT-continual\cite{TENT} & 73.07* & \underline{78.66}* & 71.94* & 46.81* & 70.20* & 68.13 \\ 
DLTTA \cite{dltta} & \underline{75.11} & 78.85* & 73.89* & 51.64* & 69.71* & 69.84 \\ 
DUA \cite{dua} & 72.28* & 76.59* & 70.13* & 56.57* & 71.38* & 69.31 \\ 
SAR \cite{sar} & 74.55* & 77.71* & 70.78* & 55.40* & 71.72* & 70.03 \\ 
DomainAdaptor\cite{domainadaptor} & 74.50* & 76.39* & 71.81* & \underline{56.78}* & 70.55* & 70.01 \\
MoASE\cite{moase} & 74.45 & 78.46& 72.24& \textbf{63.27} & 67.32 & \underline{71.15}\\
CoTTA \cite{cotta} &  75.06* & 78.09* & 73.79* & 51.69* &\underline{75.29}*& 70.78\\
VPTTA$^{\dagger}$\cite{vptta} & 74.41*& 77.47*& \textbf{74.14}* &54.03*& 67.85* & 69.58 \\
 
\hline

\textbf{MGIPT (Ours)} & \textbf{75.14} & \textbf{78.92} & \underline{73.89} & 55.89 &  \textbf{77.91} &  \textbf{72.35} \\\hline
\end{tabular}
}
\label{tab:fundus_results}
\end{table}

\begin{table*}[ht]
\caption{Performance of our method, ``Source Only'' baseline, and seven competing methods on the polyp segmentation task. The best and second-best results in each column are highlighted in \textbf{bold} and \underline{underline}, respectively. * denotes statistical significance (p\textless 0.05) using Wilcoxon signed-rank test by comparing other methods to ours. $^{\dagger}$ means removing memory bank.} 
\centering
\scalebox{0.72}{
\begin{tabular}{l|ccc|ccc|ccc|ccc|ccc}
\hline
Methods & \multicolumn{3}{c|}{Domain A} & \multicolumn{3}{c|}{Domain B} & \multicolumn{3}{c|}{Domain C} & \multicolumn{3}{c|}{Domain D} & \multicolumn{3}{c}{Average} \\ \hline
Metrics & DSC & $E_{\phi}^{\max}$ & $S_{\alpha}$ & DSC & $E_{\phi}^{\max}$ & $S_{\alpha}$ & DSC & $E_{\phi}^{\max}$ & $S_{\alpha}$ & DSC & $E_{\phi}^{\max}$ & $S_{\alpha}$ & DSC $\uparrow$ & $E_{\phi}^{\max}$ $\uparrow$ & $S_{\alpha}$ $\uparrow$ \\ \hline
Source Only (PraNet) & 79.90 & 87.97 & 84.66 & 66.33 & 78.51 & 76.72 & 73.89 & 84.64 & 81.28 & 82.95 & 90.84 & 88.08 & 75.77 & 85.49 & 82.69 \\ \hline
TENT~\cite{TENT} & 74.86* & 84.58* & 80.52* & 67.51* & 78.66* & 76.05* & 71.79* & 83.42* & 80.08* & 73.55* & 83.38* & 82.41* & 58.43 & 71.67 & 73.57 \\ 
CoTTA~\cite{cotta} & 76.46* & 85.37* & 82.56* & 66.77* & 76.75* & 79.17* & 71.39* & 83.42* & 80.08* & 73.55* & 83.38* & 82.54* & 71.33 & 81.34 & 81.11 \\ 
DLTTA\cite{dltta} & 76.27* & 85.23* & 81.20* & 66.58* & 77.00* & 79.24* & 63.72* & 78.23* & 75.56* & 71.20* & 81.32* & 83.47* & 69.44 & 80.45 & 80.17 \\ 
DUA\cite{dua} & 78.93* & 87.33* & 83.96* & 66.84* & 78.52* & 77.51* & \underline{76.53}* & \underline{86.45}* & \underline{83.08}* & \underline{86.24}* & \underline{93.23}* & \underline{89.82}* & 77.13 & 86.39 & 83.58 \\ 
SAR \cite{sar} & 76.48* & 85.69* & 81.49* & 66.45* & 77.35* & 77.05* & 71.66* & 83.06* & 80.85* & 70.41* & 81.76* & 84.45* & 71.20 & 81.65 & 80.00 \\ 
DomainAdaptor \cite{domainadaptor} & 77.48* & 86.31* & 82.47* & 70.82* & 81.76* & 80.88* & 71.96* & 83.06* & 79.97* & 76.89* & 85.89* & 84.45* & 74.29 & 84.26 & 81.93 \\ 
MoASE\cite{moase} & 70.05* & 81.75*&77.42* & 68.62*& 80.32*& 78.13* & 64.69*& 78.29*& 75.04* & 71.67*& 81.41* & 81.40* &  68.76 & 80.44 & 78.00 \\

VPTTA$^{\dagger}$ \cite{vptta}& \underline{80.91}* &  \underline{88.87}* &  \underline{84.48}* & \underline{75.51}* & \underline{87.29}*  &\underline{83.86}* & 76.09*& 86.21*& 82.85*& 85.19*& 92.13* & 88.78*& \underline{79.42} & \underline{88.62} & \underline{84.99} \\
\hline

 
\textbf{MGIPT (Ours)} & \textbf{81.40} 	&\textbf{89.17} 	&\textbf{85.11} &\textbf{76.89} & 	\textbf{87.54} &	\textbf{84.12} &	\textbf{77.56} &	\textbf{87.39} 	& \textbf{83.58} &	\textbf{86.68} 	& \textbf{93.70} &	\textbf{90.04}	&\textbf{80.64} &	\textbf{89.45} 	& \textbf{85.71} 
 \\ \hline
\end{tabular}
}

\label{tab:polyp_results}
\vspace{-.1in}
\end{table*}

\begin{table*}[t]
    \centering
    \small
    \caption{Performance of our method and five competing methods on the OD/OC segmentation task in a long-term continual test-time adaptation. The \textcolor{red}{red} numbers indicate the performance gain relative to the ``Source Only'' baseline. The performance degradation(PD) is calculated between the overall average DSC and the average DSC of round 1. ``Avg'': ``Average''.}
\label{tab:longterm}
    \scalebox{0.68}{
    \begin{tabular}{l|cccccc|cccccc|cccccc|cc}
    \toprule
    Time & \multicolumn{18}{c}{$\xrightarrow{\hspace{20cm}}$} \\
    \midrule
    Round & \multicolumn{6}{c|}{1} & \multicolumn{
6}{c|}{2} & \multicolumn{6}{c|}{3} & \multicolumn{2}{c}{} \\
    Methods & A & B & C & D & E & Ave & A & B & C & D & E & Ave & A & B & C & D & E & Avg & DSC↑ & PD.↓ \\
    \midrule
    Source Only &64.53& 76.06 &71.18 &52.67& 64.87& 65.86 &64.53 &76.06& 71.18 &52.67& 64.87& 65.86& 64.53 &76.06&71.18 &52.67& 64.87 &65.86 &65.86& - \\
TENT & 73.07 & 75.66 & 71.94 & 46.81 & 70.20 & 68.13 & 62.09 & 69.32 & 70.67 & 30.02 & 68.22 & 61.86 & 57.05 & 62.47 & 70.20 & 30.02 & 66.37 & 59.02 & 63.01*(\textcolor{red}{-2.85}) & 5.12 \\
CoTTA & 75.06 & 78.09 & 73.79 & 51.69 &75.29&70.78 & 74.31 & 75.00 & 67.09 & 51.04 & 68.28 & 67.32 & 73.22 & 74.33 & 66.72 & 50.23 & 67.08 & 66.32 & 67.84*(\textcolor{red}{+1.68}) & 1.44 \\
DLTTA & 75.11 & 75.85 & 73.89 & 51.04 & 69.71 & 69.34 & 74.14 & 79.60 & 74.25 & 45.06 & 69.04 & 68.43 & 72.38 & 78.93 & 72.27 & 42.37 & 69.36 & 67.14 & 68.47*(\textcolor{red}{+2.01}) & 1.37 \\
SAR & 74.55 & 77.71 & 70.78 & 55.40 & 71.72 & 70.03 & 74.74 & 78.09 & 71.00 & 59.13 & 69.02 & 69.00 & 74.90 & 78.24 & 71.18 & 50.16 & 68.44 & 68.58 & 69.20*(\textcolor{red}{+3.34}) & 0.83 \\
MoASE & 74.45 & 78.46& 72.24& 63.27 & 67.32 & 71.15 & 73.72&77.88 &71.33   &62.93& 66.58& 70.48 & 72.81 &76.98& 71.05&  62.78 & 65.28 & 69.78 & \underline{70.47}*(\textcolor{red}{+4.61}) & 0.68
 \\
VPTTA$^{\dagger}$  & 74.41& 77.47& 74.14 &54.03& 67.85 & 69.58 &  73.51& 76.43 &73.72 & 53.18&  66.78 & 68.72 & 72.91  &75.39 & 72.52 &53.52 &66.75 & 68.22 & 68.84*(\textcolor{red}{+2.68}) & 0.74 \\
\textbf{Ours} &75.14& 78.92 &73.89 &55.89& 77.91 &\textbf{72.35}&  
74.98& 78.91& 73.71 &55.80& 77.88 &\textbf{72.26} & 
74.94& 78.88& 73.63& 55.79& 77.94& \textbf{72.24} & \textbf{72.28}(\textcolor{red}{\textbf{+6.42}}) & \textbf{0.07}\\
    \bottomrule
    \end{tabular}
    }
    \vspace{-.05in}
\end{table*}

\subsection{Ablation Study}
We conducted comprehensive ablation studies on the OD/OC segmentation task to evaluate the effectiveness of each proposed component: instance-level prompt (IP), global-level prompt (GP), and BN statistics calibration (BN Calib.). As shown in Table~\ref{tab:ablation-study}, we derive several key findings.
(1) Effectiveness of BN Calib.: applying BN statistics calibration consistently boosts performance (Row \#2), demonstrating its importance across configurations.
(2) Limited improvement of single-scale GP: utilizing a single fixed-scale global prompt yields modest improvements (Row \#3), indicating the limitations of relying solely on domain-level knowledge transfer. 
(3) Enhancements with EMA and MS: the effectiveness of GP is enhanced through exponential moving average (EMA) (Row \#4) and multi-scale (MS) (Row \#5) mechanisms, capturing diverse multi-frequency style changes. 
(4) Limitations of single-scale IP (Fix): a single-scale IP shows limited improvement when used alongside GP (Row \#5 vs. \#6). 
(5) Improvements with early stopping (ES) and progressive prompt tuning (PPT): ES and PPT strategies enhance IP with optimal instance scale selection (Row \#7 and \#8). These enhancements lead to consistent performance gains.
(6) Complementary nature of IP and GP: The combination of IP and GP demonstrates significant performance improvements (Row \#2, \#5 and \#8), highlighting their complementary strengths. 

\subsection{Comparison with State-of-the-Art Methods}
 In Table~\ref{tab:polyp_results}, we evaluate our MGIPT framework against the ``Source Only'' baseline (training on source domain and testing without adaptation) and seven state-of-the-art methods, including two pseudo-label based method (CoTTA~\cite{cotta} and MoASE~\cite{moase}), two entropy-based methods (TENT-continual~\cite{TENT} and SAR~\cite{sar}), a method dynamically adjusting the learning rate (DLTTA ~\cite{dltta}), a method combining entropy-loss and BN statistics fusion (DomainAdaptor~\cite{domainadaptor}), a method using visual prompts removing its memory bank~\cite{vptta}, and a BN statistics-based method (DUA~\cite{dua}). Notably, in real biomedical applications of test time, patient data is strictly protected, making it infeasible to retain a memory bank containing previously seen images. Therefore, we remove the memory bank in VPTTA for a fair comparison. All methods were re-implemented under the same experimental setup.
 

\par
\begin{figure}[htbp]
    \centering
     \includegraphics[width=0.9\linewidth]{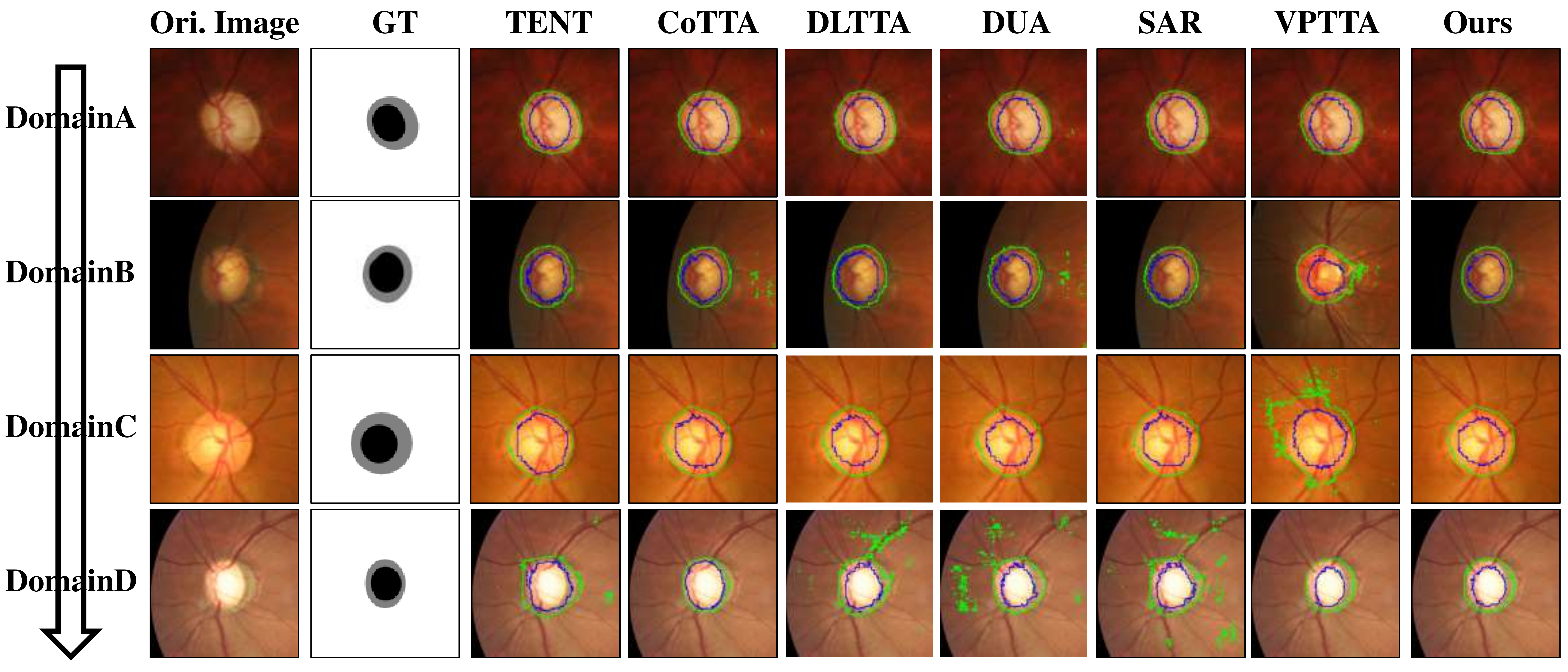}
\caption{Qualitative comparison of segmentation results on target domains for OD/OC benchmark. ‘Ori’: ‘Original’.}
\label{fig-segmentation}
\end{figure}

\noindent\textbf{Results on OD/OC Segmentation Task.} 
The results on the OD/OC segmentation task are presented in Table~\ref{tab:fundus_results} and Fig.~\ref{fig-segmentation}.
For each task, we use the pretrained source model and evaluate it on the remaining domains (target domains), e.g. ``A'' in Table~\ref{tab:fundus_results} means using Domain A as source domain and the left domains as target. All competing methods overall outperform the ``Source Only" baseline, demonstrating their effectiveness in addressing domain shifts. Notably, our MGIPT method achieves the best average performance across all domains, as well as the highest score when the source domain is A, B and E. This highlights the robustness and superior adaptability of our method to varying domain distributions.

\par\noindent\textbf{Results on Polyp Segmentation Task.} 
We conducted comparison experiments on the polyp segmentation task under the same experimental setup, as reported in Table~\ref{tab:polyp_results}. Unlike the results from the OD/OC segmentation task, model updating methods (e.g., TENT-continual, MoASE, CoTTA, DLTTA, SAR, and DomainAdaptor) perform notably worse, even inferior to the ``Source Only'' baseline. 
This is because polyps are relatively hidden, so under distribution shifts models tend to segment almost nothing instead of producing poor segmentation. This behavior leads to confident yet incorrect low-entropy predictions that generate misleading gradients for these methods.
The backward-free DUA method maintains competitive performance due to its reliance on batch normalization statistics. But it struggles with Domain A as the source, showing its limitations in test-time adaptation without additional training on the current test sample. MGIPT achieves the best overall performance across all domains and metrics, demonstrating its adaptability and effectiveness in addressing distribution shifts.

\begin{figure}[htbp]
    \centering
    \includegraphics[width=0.95\linewidth]{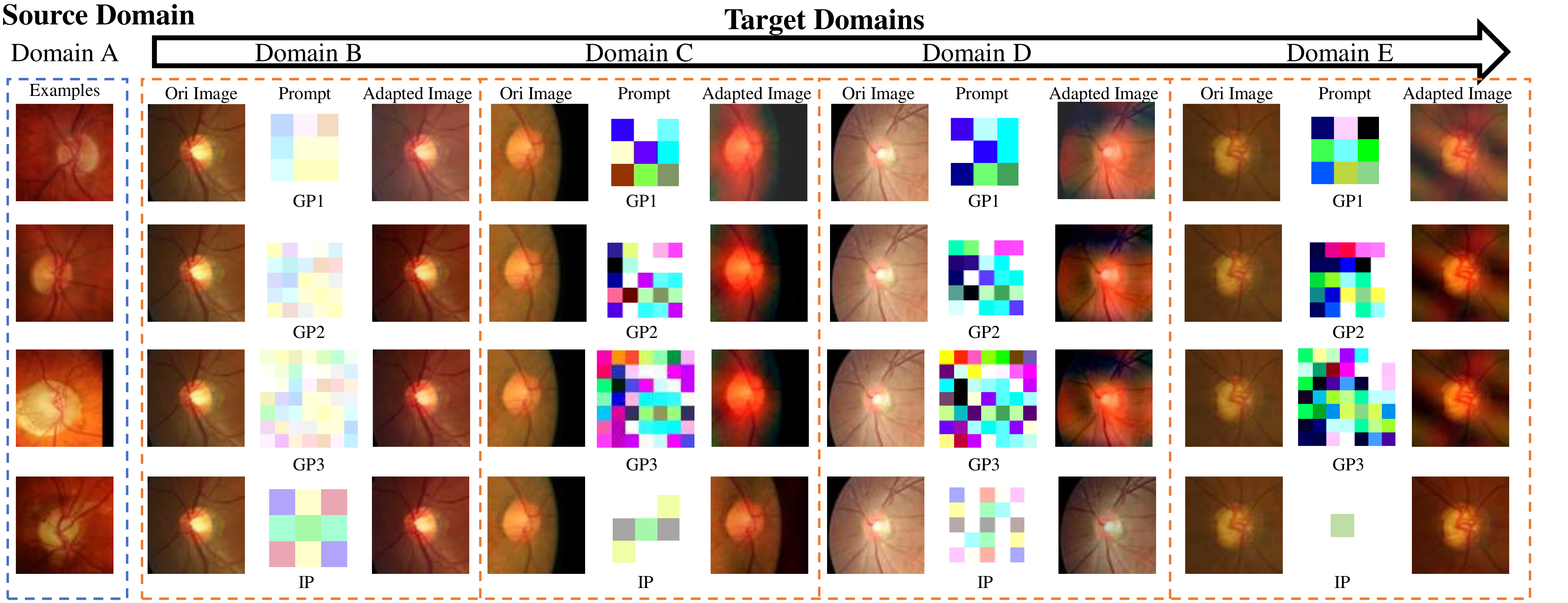}
    \vspace
{-5pt}
    \caption{Visualization of the original images, finetuned prompts, and adapted images on the OD/OC segmentation task. We normalize the prompts to [0, 1] for better visualization. We also show some examples of source domain on the upper side of this diagram. ``Ori'': Abbreviation of ``Original''.}
    \label{fig3}
    \vspace{-.15in}
\end{figure}

\par\noindent\textbf{Results on Long-Term CTTA.}
To evaluate our method in long-term CTTA~\cite{cotta}, we conducted experiments on the OD/OC segmentation task over multiple rounds (i.e. 1, 2, 3) and reported the results in Table~\ref{tab:longterm}. Following long-term setup in VPTTA~\cite{vptta}, we compared our method with TENT, CoTTA, MoASE, DLTTA, SAR, and VPTTA, the first five of which involve continuous training of the source model.  
The performance across different rounds, using the same source domain, reflects the extent of catastrophic forgetting, while the average performance over multiple rounds highlights error accumulation across distinct domains. The results reveal that methods optimizing loss functions (e.g., CoTTA, DLTTA, SAR), resetting the model (e.g., CoTTA, MoASE, SAR) or maintaining a memory bank (e.g., VPTTA) can mitigate error accumulation and catastrophic forgetting. However, TENT, which utilizes a vanilla entropy-minimization loss, suffers from significant performance degradation. Although VPTTA demonstrates improvements in addressing catastrophic forgetting and error accumulation, its reliance on fixed-size global prompts with memory banks limits its ability to capture optimal instance-specific knowledge and multi-scale diversity. Furthermore, the use of a memory bank of previous prompts results in error accumulation due to the fixed-scale prompts, which becomes more pronounced over long term as noise compounds. 
In contrast, our method maintains superior performance even after three rounds of testing, with minimal performance degradation. This underscores the effectiveness of training dual-level prompts, which balance global and instance-specific knowledge, outperforming methods that rely solely on updating the pre-trained model or fixed-size prompts.

\begin{figure}[htbp]
  \centering
   \includegraphics[width=0.75\linewidth]{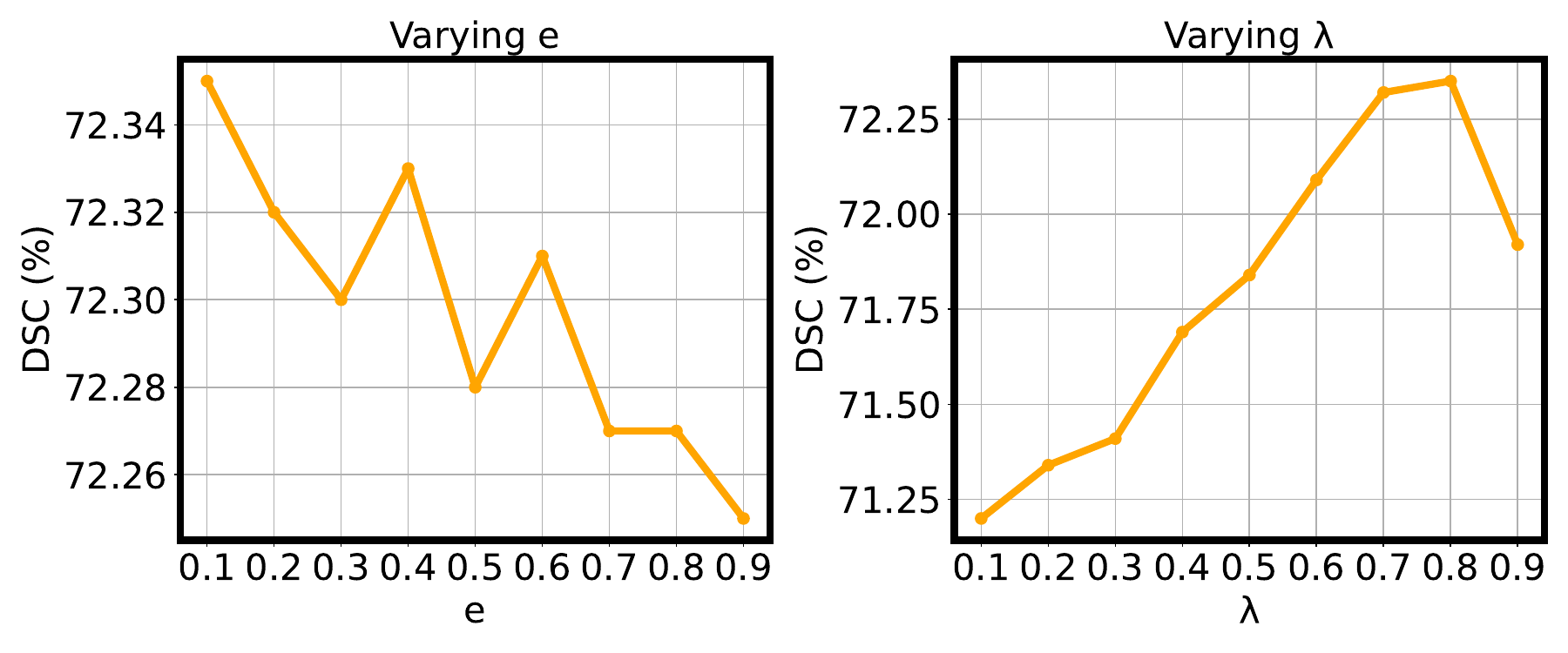}
   \vspace{-6pt}
   \caption{Performance with various $e$ and $\lambda$ on OD/OC task.}
   \label{fighyper}
   \vspace{-.2in}
\end{figure}

\subsection{Further Investigation}
\par\noindent\textbf{Analysis of Hyperparameter $e$ and $\lambda$.}
We conducted extensive experiments on the OD/OC segmentation task to analyze the sensitivity of our method to the two key hyperparameters: the EMA decay rate $e$ and the BN calibration weight $\lambda$. As shown in Fig.~\ref{fighyper}, while the performance remains relatively stable across different values of $e$, we observe optimal results at $e=0.1$. The limited variance in performance (±0.1\% DSC) suggests that our method is robust to this hyperparameter.
Fig.~\ref{fighyper} also shows the impact of BN calibration weight $\lambda$ on model performance. As $\lambda$ increases from 0.1 to 0.8, we observe a consistent improvement, while excessively large $\lambda$ leads to performance degradation, likely due to over-reliance on source domain statistics. 
Through hyperparameter analysis on OD/OC dataset, we set $e=0.1$ and $\lambda=0.8$ as default configuration across all experimental settings, showing consistent improvement over baselines on all tasks.

\par\noindent\textbf{Visualization.}
We visualized the fine-tuned prompts and adapted images on the OD/OC segmentation task in Fig.~\ref{fig3} and displayed the DSC values predicted on the original and adapted images below the corresponding images. It can be observed that:
(1) Multi-scale global prompts (GP) are generally similar, while instance prompts can be different with GPs, showing IPs and GPs extract different knowledge; 
(2) The appearances of adapted images are closer to the source domain, demonstrating the reduction of distribution shifts via our method.

\section{Conclusion}
In this paper, we introduced MGIPT, a novel dual-prompt method for continual test-time adaptation, addressing the critical challenges of error accumulation and catastrophic forgetting. By leveraging complementary adaptive gradual instance prompts and multi-scale global prompts, MGIPT adapts effectively to instance-specific and domain distribution shifts without altering the pre-trained model. The effectiveness of MGIPT lies in its dual-prompt strategy, which adaptively aligns test images through instance-specific and global prompts with lightweight parameter updates.
Our experiments on multiple medical image segmentation benchmarks demonstrate that MGIPT outperforms existing methods, particularly in long-term adaptation scenarios. 
Our limitation is that MGIPT requires iterative adaptation to each target sample to find the individual optimal prompt size, resulting in increased computational time. 





\section*{Acknowledgment}
This work was funded by the National Natural Science Foundation of China (No.62572166).


\end{document}